\documentclass{article}
\usepackage{amsmath}
\usepackage{geometry}
\usepackage{hyperref}
\usepackage{graphicx}
\usepackage{float}
\usepackage{amsmath}
\usepackage{amsfonts}
\usepackage{amssymb}

\title{\textbf{Demonstrating the Continual Learning Capabilities and Practical Application of Discrete-Time Active Inference}}
\author{}
\date{}

\begin{document}

\maketitle
\thispagestyle{empty}

\vspace{1cm}

\begin{center}
    \textbf{Rithvik Prakki} \\
    University of North Carolina at Chapel Hill \\
    \texttt{rprakki@unc.edu}
\end{center}

\vspace{1cm}

\begin{abstract}
Active inference provides a powerful mathematical framework for understanding how agents—biological or artificial—interact with their environments, enabling continual adaptation and decision-making. It combines the principles of Bayesian inference and free energy minimization to model the process of perception, action, and learning in uncertain and dynamic contexts. Unlike reinforcement learning, active inference integrates both exploration and exploitation seamlessly, driven by a unified objective to minimize expected free energy. In this paper, we present a  continual learning framework for agents operating in discrete time environments, using active inference as the foundation. We derive the core mathematical formulations of variational and expected free energy and apply these principles to the design of a self-learning research agent. This agent continually updates its beliefs and adapts its actions based on new data, without manual intervention. Through experiments in dynamically changing environments, we demonstrate the agent's ability to relearn and refine its internal models efficiently, making it highly suitable for complex and volatile domains such as quantitative finance and healthcare. We conclude by discussing how the proposed framework generalizes to other systems and domains, positioning active inference as a robust and flexible approach for adaptive artificial intelligence.
\end{abstract}

\newpage

\section{Introduction}

The \textit{Free Energy Principle (FEP)}, proposed by Friston, provides a unifying computational framework that integrates learning, perception, action, and decision-making. This principle posits that biological systems (or artificial agents) maintain themselves in their characteristic states by minimizing the difference between the predictions of their internal model and their actual sensory data, a quantity known as free energy. In essence, FEP suggests that living systems are driven by the need to reduce surprise in their interactions with the environment, formalized through the minimization of variational free energy.

\subsection{Active Inference and Free Energy Principle}

At the core of \textit{Active Inference} (AIF), which operationalizes the Free Energy Principle, are two key objective functions:
\begin{itemize}
    \item \textbf{Variational Free Energy (VFE)}: This function quantifies the fitness of an agent's internal model concerning the sensory observations it receives from its environment by mapping latent (hidden) states to sensory outcomes. The agent minimizes this free energy to maintain coherence between its beliefs about the world and incoming sensory evidence.
    \item \textbf{Expected Free Energy (EFE)}: This function governs the agent’s policy selection by combining extrinsic value (goal-oriented behavior, aligned with the agent's prior preferences) and epistemic value (exploration-driven behavior to reduce uncertainty in the agent's model of the world).
\end{itemize}

Discrete-Time Active Inference is framed in the context of \textit{Partially Observable Markov Decision Processes (POMDPs)}, where agents use sensory and proprioceptive information to form probabilistic beliefs about the hidden (latent) states of the world.

\subsection{Mathematical Derivation of Variational Free Energy (VFE)}

Variational Free Energy (VFE) is a quantity that measures the dissimilarity between an agent’s internal model of the world and the real-world sensory data it receives. It is used to approximate Bayesian inference in scenarios where calculating exact posteriors is computationally intractable. To derive VFE, we begin with Bayes’ rule:

\begin{equation}
P(s | o) = \frac{P(o | s)P(s)}{P(o)}
\end{equation}

Here, \( P(s | o) \) is the posterior distribution over hidden states \( s \) given observations \( o \), \( P(o | s) \) is the likelihood (i.e., how likely the observations are given the hidden states), \( P(s) \) is the prior distribution over hidden states, and \( P(o) \) is the marginal likelihood, also known as the evidence.

In active inference, the agent cannot compute this posterior \( P(s | o) \) directly, as the marginal likelihood \( P(o) \) is intractable due to the high-dimensional nature of real-world data. Instead, the agent approximates the posterior with a simpler distribution \( Q(s) \), known as the variational distribution, which the agent iteratively improves. The agent minimizes the difference between the true posterior \( P(s | o) \) and the variational approximation \( Q(s) \) using the Kullback-Leibler (KL) divergence:

\begin{equation}
D_{\mathrm{KL}}[Q(s) \| P(s | o)] = \int Q(s) \ln \frac{Q(s)}{P(s | o)} ds
\end{equation}

Since the true posterior \( P(s | o) \) is unknown, we aim to minimize this divergence indirectly by minimizing the \textit{free energy} \( F \), which is an upper bound on the negative log-evidence, \( -\ln P(o) \). To express this in terms of known quantities, we substitute Bayes’ rule into the expression for the KL divergence:

\begin{equation}
D_{\mathrm{KL}}[Q(s) \| P(s | o)] = \int Q(s) \ln \frac{Q(s)}{P(o | s) P(s) / P(o)} ds
\end{equation}

This can be split into three terms:

\begin{equation}
D_{\mathrm{KL}}[Q(s) \| P(s | o)] = \int Q(s) \ln \frac{Q(s)}{P(s)} ds - \int Q(s) \ln P(o | s) ds + \ln P(o)
\end{equation}

Thus, the KL divergence becomes:

\begin{equation}
D_{\mathrm{KL}}[Q(s) \| P(s | o)] = D_{\mathrm{KL}}[Q(s) \| P(s)] - \mathbb{E}_{Q(s)}[\ln P(o | s)] + \ln P(o)
\end{equation}

In this expression, \( D_{\mathrm{KL}}[Q(s) \| P(s)] \) is the KL divergence between the approximate posterior \( Q(s) \) and the prior \( P(s) \), which we call the \textbf{complexity} term. The second term \( \mathbb{E}_{Q(s)}[\ln P(o | s)] \) is the expected log-likelihood of the observations under the approximate posterior, known as the \textbf{accuracy} term. 

The final term, \( \ln P(o) \), is the log-evidence (also known as the marginal likelihood), which does not depend on \( Q(s) \) and is therefore a constant with respect to the minimization of free energy.

Since \( \ln P(o) \) is independent of \( Q(s) \), minimizing the KL divergence \( D_{\mathrm{KL}}[Q(s) \| P(s | o)] \) is equivalent to minimizing the \textit{variational free energy} \( F \), which we define as:

\begin{equation}
F[Q(s)] = D_{\mathrm{KL}}[Q(s) \| P(s)] - \mathbb{E}_{Q(s)}[\ln P(o | s)]
\end{equation}

This free energy consists of two competing terms:
\begin{itemize}
    \item \textbf{Complexity:} \( D_{\mathrm{KL}}[Q(s) \| P(s)] \), which penalizes divergence between the approximate posterior and the prior. Lowering this term encourages simplicity in the model, favoring posteriors \( Q(s) \) that are close to the prior \( P(s) \).
    \item \textbf{Accuracy:} \( \mathbb{E}_{Q(s)}[\ln P(o | s)] \), which measures how well the model’s predictions \( P(o | s) \) fit the observed data \( o \). Maximizing this term ensures that the model is accurately predicting sensory observations.
\end{itemize}

The free energy \( F \) can be minimized by iteratively updating the approximate posterior \( Q(s) \), driving the agent to strike a balance between complexity (favoring simpler models) and accuracy (favoring models that predict observations well).

\subsubsection{Expected Free Energy (EFE)}

While Variational Free Energy (VFE) focuses on inferring hidden states based on current observations, Expected Free Energy (EFE) extends this principle into the future by evaluating potential actions (or policies) that an agent can take. Just as VFE is used to update the agent’s beliefs about the hidden states \( s \) given observations \( o \), EFE evaluates the likely outcomes of different policies \( \pi \), guiding the agent to select actions that minimize future free energy.

Similar to VFE, the derivation of EFE follows from Bayes' rule. However, instead of inferring hidden states based solely on observations, we now consider how policies \( \pi \) influence both the hidden states and the observations. The posterior over hidden states and policies is given by:

\begin{equation}
P(s | o, \pi) = \frac{P(o | s, \pi) P(s, \pi)}{P(o, \pi)}
\end{equation}

Here, \( P(s | o, \pi) \) is the posterior over hidden states given observations and policies, \( P(o | s, \pi) \) is the likelihood of the observations conditioned on both the hidden states and the policy, \( P(s, \pi) \) is the joint prior over states and policies, and \( P(o, \pi) \) is the marginal likelihood of observations and policies.

From this starting point, we can derive the expected free energy, \( G_\pi \), following a similar process as we used for VFE. The goal of EFE is to select policies that minimize the free energy expected over future observations, balancing the agent’s desire to gather information (epistemic value) and achieve its goals (extrinsic value).

\subsection{Learning in Active Inference: The Dirichlet-Categorical Model}

In Active Inference, learning occurs by updating the generative model's parameters based on new observations. This process is typically modeled with a \textit{Dirichlet-Categorical} framework, where Bayesian inference is performed using a categorical distribution as the likelihood and a Dirichlet distribution as the prior.

A \textbf{Dirichlet distribution}, denoted by \( \text{Dir}(\theta|\alpha) \), is defined over probability vectors and serves as the conjugate prior for the categorical distribution. This allows for sequential updating of beliefs as new data is observed. The Dirichlet distribution is expressed as:

\begin{equation}
p(\theta | \alpha) = \frac{\Gamma\left( \sum_{k=1}^{K} \alpha_k \right)}{\prod_{k=1}^{K} \Gamma(\alpha_k)} \prod_{k=1}^{K} \theta_k^{\alpha_k - 1}
\end{equation}

Where \( \alpha = (\alpha_1, \alpha_2, \dots, \alpha_K) \) are concentration parameters, and \( \Gamma(\cdot) \) is the gamma function. The parameters \( \alpha_k \) represent prior counts that encode the agent’s confidence in each possible outcome before observing new data.

As the agent receives new observations, these concentration parameters are updated through a process analogous to "counting" the co-occurrences of certain states and observations over time. The posterior distribution over parameters \( \theta \), given new observations \( x \), becomes:

\begin{equation}
p(\theta | x, \alpha) \propto \text{Dir}(\theta | \alpha + x)
\end{equation}

This form of learning enables the agent to accumulate evidence over time, updating its beliefs about state-outcome mappings in a flexible and adaptive manner.

\subsection{Learning the A Matrix}

A crucial part of the generative model in Active Inference is the \textbf{A matrix}, which encodes the likelihood mapping from hidden states \( s \) to observations \( o \). Learning the A matrix is central to the agent’s ability to adapt to its environment, as it allows the agent to refine its understanding of how hidden states generate sensory data.

The update rule for the A matrix is expressed as:

\begin{equation}
a_{t+1} = \omega \cdot a_t + \eta \cdot \sum_{\tau} o_\tau \otimes s_\tau
\end{equation}

Where:
\begin{itemize}
    \item \( a_t \) are the concentration parameters for the A matrix at time \( t \),
    \item \( o_\tau \) is the observation at time \( \tau \),
    \item \( s_\tau \) is the inferred hidden state at time \( \tau \),
    \item \( \eta \) is the learning rate that controls the adaptation speed, and
    \item \( \omega \) is the forgetting rate that controls the extent to which past observations are discounted.
\end{itemize}

This Hebbian-like learning rule updates the A matrix by strengthening associations between states and observations that co-occur. For example, if the agent observes \( o = [1, 0, 0]^T \) while in state \( s = [1, 0]^T \), the association between state 1 and observation 1 is reinforced. This allows the agent to build stronger beliefs about the causal structure of the environment.

In addition, the process of policy selection relies on the minimization of \textit{Expected Free Energy (EFE)}, which includes a \textbf{novelty-seeking} term that drives exploration. This term quantifies the expected change in beliefs as a result of receiving a new observation. The EFE in this context is:

\begin{equation}
G_\pi = D_{\mathrm{KL}}\left[q(o_\tau|\pi) \| p(o_\tau)\right] + \mathbb{E}_{q(s_\tau|\pi)}\left[H[p(o_\tau|s_\tau)]\right] - \mathbb{E}_{p(o_\tau|s_\tau) q(s_\tau|\pi)}\left[D_{\mathrm{KL}}\left[q(A|o_\tau, s_\tau) \| q(A)\right]\right]
\end{equation}

The third term drives the agent to seek novel observations that maximize the difference between prior and posterior beliefs about the mappings in the A matrix. By doing so, the agent can actively explore and refine its internal model, improving its understanding of how states generate observations.

\subsection{Application to Self-Learning Agents}

In this paper, we extend this framework to the design of \textbf{self-learning agents} capable of continually updating their generative models to adapt to changing environments. Specifically, we focus on the learning of the A matrix, as it is essential for the agent’s ability to infer the dynamics of its environment. By using Active Inference and Dirichlet-Categorical models, we propose a mechanism for integrating continual learning into agentic systems, allowing agents to optimize their performance over time while responding to novel observations.

This ability to learn continually over time ensures that the agent can interact with dynamic environments without requiring constant updates or interventions by a human operator. This capability is especially relevant in domains such as quantitative finance, where strategies can become obsolete over time due to market fluctuations, or in research, where resources constantly evolve in their relevance. In these scenarios, Active Inference provides an abstraction layer that continuously tunes the agent’s performance, thus adapting to new challenges in real-time.

\section{Active Inference as an Agentic Learning Mechanism}

\subsection{Motivating the Problem}

For every aspect of an agentic workflow, there are elements that need to be optimized over time. Consider an agent that operates within a dynamic environment, where the relationships between states, observations, and actions continuously evolve. If the system lacks the ability to adapt to these changes, its performance will degrade. Traditional methods require manual intervention, continual updates, or predefined structures that can become obsolete.

However, with Active Inference, by simply defining the environment and creating pathways for the agent to act on and receive observations from the environment, we can build agents that are self-sustaining and highly adaptable over time. Active Inference agents can automatically learn new structures and processes as needed without the need for external updates, making them resilient to change. 

\subsection{Developing the Framework}

This section develops the logic behind the active inference framework used for implementing self-learning within any agentic workflow. While the following discussion focuses on an example research agent, the framework can be generalized to other systems and domains.

\subsubsection{Definitions}

Operating in discrete time, a Partially Observable Markov Decision Process (POMDP) models states, observations, policies, and all inputs and outputs of Free Energy. In this framework, the \textbf{A matrix} represents the likelihood mapping from hidden states to observations. The state factors and observation modalities are learned over time rather than pre-specified, allowing the agent to adapt as new data becomes available.

For example, in a research process, the state factors may represent different research methods, while the observation modalities represent the usefulness of a research output. Over time, the system refines its understanding of which research methods yield the most useful results in a given context.

\subsubsection{Generative Model}

In this framework, the generative model is hierarchical, structured across two levels. This hierarchy is crucial for providing context to the agent, enabling it to understand both the environment it is interacting with and the specific processes within that environment.

The \textbf{top level} of the generative model consists of one state factor: the \textit{industry}, and one observation modality: the \textit{industry cue}. The purpose of this level is to provide the agent with an understanding of the high-level context—essentially, which industry it is interacting with at a given time. The agent observes an industry cue that informs it about the current industry in which the research processes are unfolding. This ensures that the agent can interpret all subsequent observations with the knowledge of the specific industry it is dealing with.

The \textbf{bottom level} of the generative model consists of two state factors: the \textit{industry} and the \textit{research process}, with three observation modalities:
\begin{itemize}
    \item \textbf{Research process}: Observes which research process is being used.
    \item \textbf{Outcome}: Observes the result of the research process in a given industry (i.e., a mapping between the research process and industry).
    \item \textbf{Industry cue}: Observes the industry cue again, passed down from the top level, to maintain awareness of the industry context.
\end{itemize}

Thus, the agent is provided with both state and observation information from the top level (industry and its cue), which is passed down to the bottom level to act as a continuous context during its exploration of research processes. When the agent observes data from the research process, it knows which industry these observations are tied to, based on the industry cue provided at the top level. The observation of the research process itself allows the agent to recognize which process is being applied, while the \textbf{outcome} provides information about the result of applying that research process in the given industry.

The \textbf{outcome} in this simplified model is a preset simulation that maps each research process to a specific result within the industry. For example, in Industry 1, Research Process 1 may always yield an "Excellent" result, while Research Process 2 may yield a "Good" result, and so on. In more practical applications, the outcome would be determined directly by interacting with the environment, but in this simplified model, the outcome serves as a simulation to help the agent learn these mappings.

By structuring the generative model in this hierarchical way, the agent is capable of maintaining awareness of its industry context while refining its understanding of how different research processes perform within that context. The hierarchical POMDP ensures that observations about industry cues are continuously processed, providing the necessary context for learning the dynamic relationships between research processes and their outcomes.

\begin{figure}[H]
    \centering
    \includegraphics[scale=0.5]{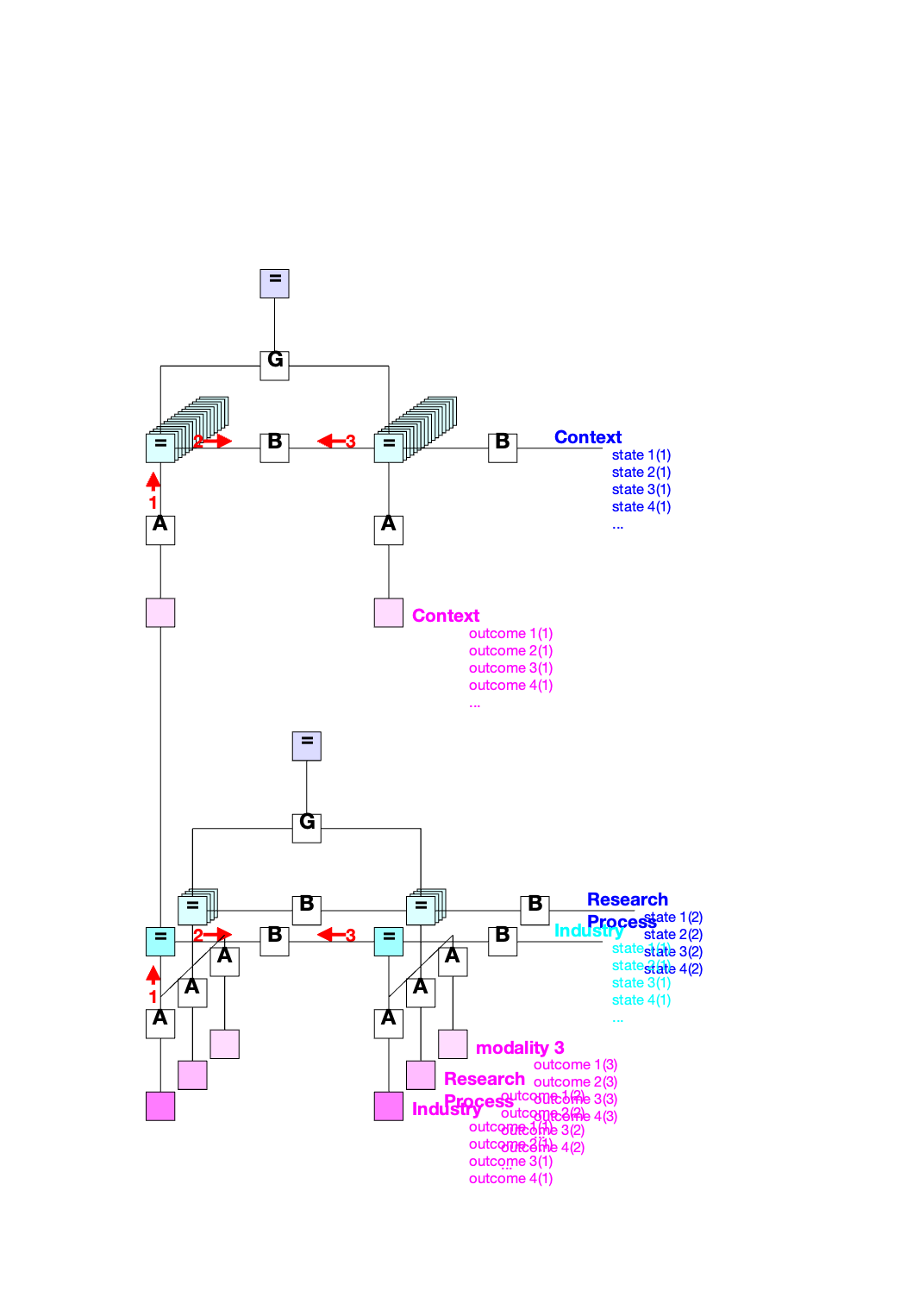} 
    \caption{Hierarchical POMDP showing research methods and their outcomes. The agent adapts to changes in the research environment by updating the state and observation factors. This structure allows for continual learning as new information becomes available.}
    \label{fig:hierarchical_pomdp}
\end{figure}

\subsubsection{Generative Process}

The generative process in this model simulates the environment that the agent interacts with. In an ideal scenario, the generative process would involve real-world data being generated by the environment in response to the agent’s actions. However, for the purpose of this research model, the generative process is simplified and simulated based on the state-observation mappings present in the \textbf{A matrix}.

In this setup, the generative process takes the agent’s policy selections and uses them to sample observations from the predefined mappings in the \textbf{A matrix}. Specifically, after the agent selects a policy (which represents its action in terms of selecting a research process), the generative process samples from the A matrix to determine the corresponding outcome for the given research process in the current industry. The A matrix contains the probabilities that govern how observations (i.e., outcomes) are generated based on the agent’s belief about the current state (industry and research process).

For example, if the agent selects Research Process 1 in Industry 1, the A matrix may indicate that there is a high probability that the outcome will be "Excellent." The generative process samples from this probability distribution, providing the agent with an observation of "Excellent." The agent then uses this observation to update its beliefs about the mappings between research processes and outcomes in the current industry.

The generative process is essentially a simulation of how the environment might behave, providing feedback to the agent about its actions. The agent is tasked with learning the structure of the environment by minimizing \textbf{Expected Free Energy (EFE)}. This involves selecting policies that are expected to reduce the agent’s uncertainty about the state-observation mappings (i.e., learning the A matrix).

Because the agent is providing observations to itself via the sampling of the A matrix, the process of learning becomes one of continuous refinement. Over time, the agent uses the observations it generates (from its interaction with the environment) to improve its internal model, refining its understanding of how different research processes produce outcomes in different industries. This allows the agent to optimize its performance, selecting policies that are expected to maximize the accuracy of its predictions about the environment.

Thus, the generative process in this framework is not only responsible for generating observations but also for driving the agent’s learning by simulating feedback loops between the agent’s policies and the environment it is modeling.

\section{Results}

To demonstrate the continual learning capabilities of the active inference agent, we tested it in two different environments: the first with predefined state-outcome mappings and the second with modified mappings for certain industries to simulate a shift in the environment. We used a scoring mechanism based on the agent's confidence in the correct outcomes, which is calculated from the belief values assigned to each possible outcome in the lowercase 'a' matrix. This matrix represents the agent’s internal belief mappings between states (industries and research processes) and observations (outcomes).

\subsection{Scoring Mechanism}

To evaluate the agent’s learning performance, we define a scoring mechanism that reflects both the accuracy and confidence of the agent in its internal belief state. The agent’s belief about the relationship between the hidden states (e.g., industries and research processes) and observations (e.g., outcomes) is represented by the lowercase 'a' matrix, denoted by \( a_{ijk} \), where:

\begin{itemize}
    \item \( i \) indexes the possible outcomes (e.g., \textit{Excellent}, \textit{Good}, etc.),
    \item \( j \) indexes the hidden states (e.g., industry),
    \item \( k \) indexes the research processes.
\end{itemize}

For each combination of hidden state \( j \) and research process \( k \), the agent assigns a probability \( a_{ijk} \) to each outcome \( i \), which reflects the agent’s belief in the likelihood of that outcome.

\subsubsection{Score Calculation}

The score for each state-process pair \( (j, k) \) is calculated based on the agent's confidence in the correct outcome relative to its confidence in the incorrect outcomes. Let \( i^* \) denote the index of the correct outcome for the given pair \( (j, k) \), and let \( i_{\max} \) denote the index of the incorrect outcome with the highest belief value (i.e., the highest \( a_{ijk} \) where \( i \neq i^* \)).

The score \( S_{jk} \) for the pair \( (j, k) \) is defined as the difference between the agent’s belief in the correct outcome and the highest belief in any incorrect outcome:

\[
S_{jk} = a_{i^*jk} - \max_{i \neq i^*} a_{ijk}.
\]

Thus, if the agent has a high belief in the correct outcome \( a_{i^*jk} \) and low belief in the incorrect outcomes, \( S_{jk} \) will be positive. Conversely, if the agent incorrectly assigns higher belief to an incorrect outcome, the score will be negative.

\subsubsection{Total Score for an Iteration}

The total score for an iteration \( t \), denoted \( S^{(t)} \), is calculated as the sum of normalized scores across all industries \( j \) and research processes \( k \):

\[
S^{(t)} = \sum_{j=1}^{16} \sum_{k=1}^{4} S_{jk}^{\text{norm}}.
\]

This total score reflects the overall performance of the agent in the current iteration, taking into account how well the agent has learned the correct mappings for each industry and research process.

\subsubsection{Negative Scores}

Negative scores occur when the agent’s belief in the correct outcome is lower than the belief in an incorrect outcome. For example, if the belief values for \textit{Excellent} and \textit{Good} were reversed:

\[
a_{11k} = 0.2, \quad a_{21k} = 5, \quad a_{31k} = 0.1, \quad a_{41k} = 0.1, \quad a_{51k} = 0.05,
\]

then the score would be:

\[
S_{1k} = 0.2 - 5 = -4.8,
\]

indicating the agent has mistakenly assigned much higher confidence to the incorrect outcome.

\subsubsection{Overall Evaluation}

By calculating and summing the scores across all industries and research processes, we obtain a measure of the agent's overall learning performance and how well it adapts to changes in the environment. This scoring mechanism allows us to assess both the accuracy and confidence of the agent’s beliefs about the environment.

\subsection{First Environment: Predefined Outcomes}

\begin{figure}[H]
    \centering
    \includegraphics[width=0.6\textwidth]{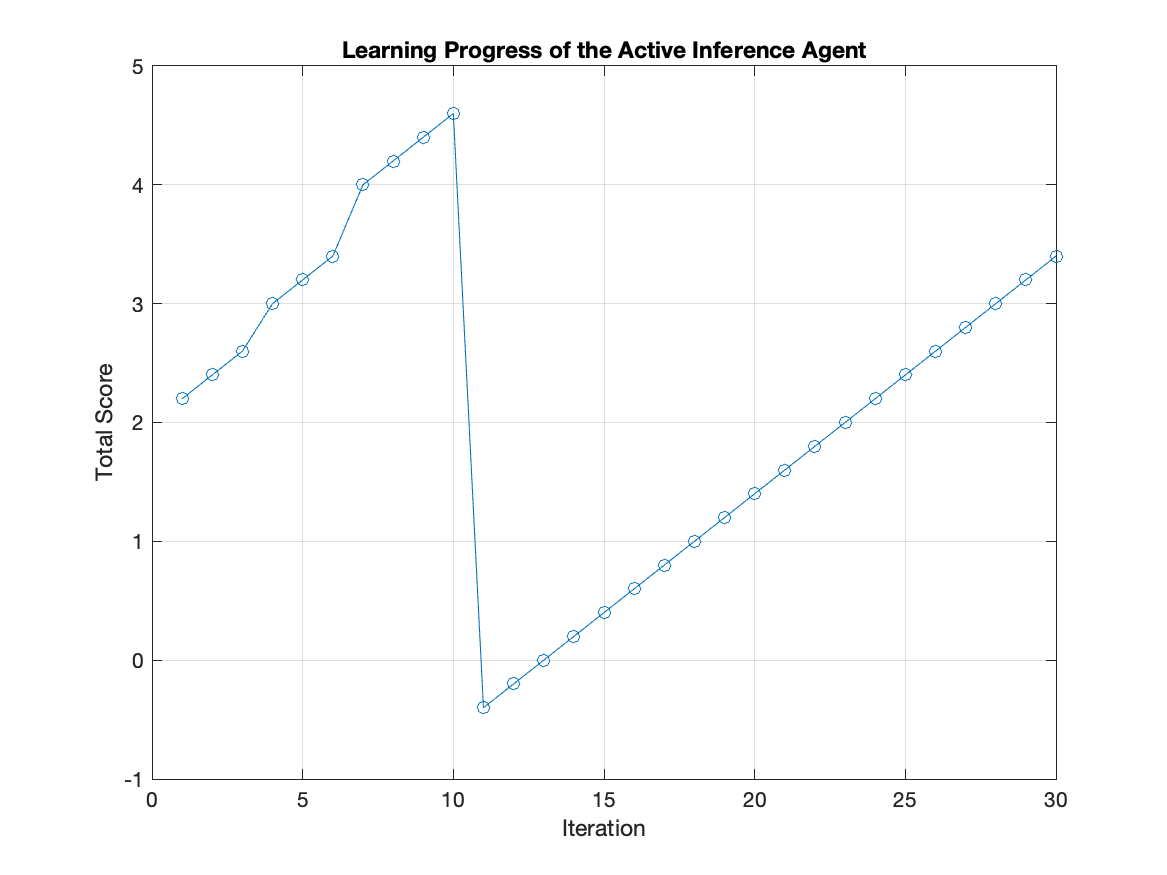}
    \caption{The results of the first 10 trials in the original environment. The agent quickly learns the dynamics of the environment, reaching a high score for Industry 1 within six iterations, demonstrating that the active inference agent is able to learn environmental dynamics effectively.}
    \label{fig:learning_progress}
\end{figure}

\begin{figure}[H]
    \centering
    \includegraphics[width=0.6\textwidth]{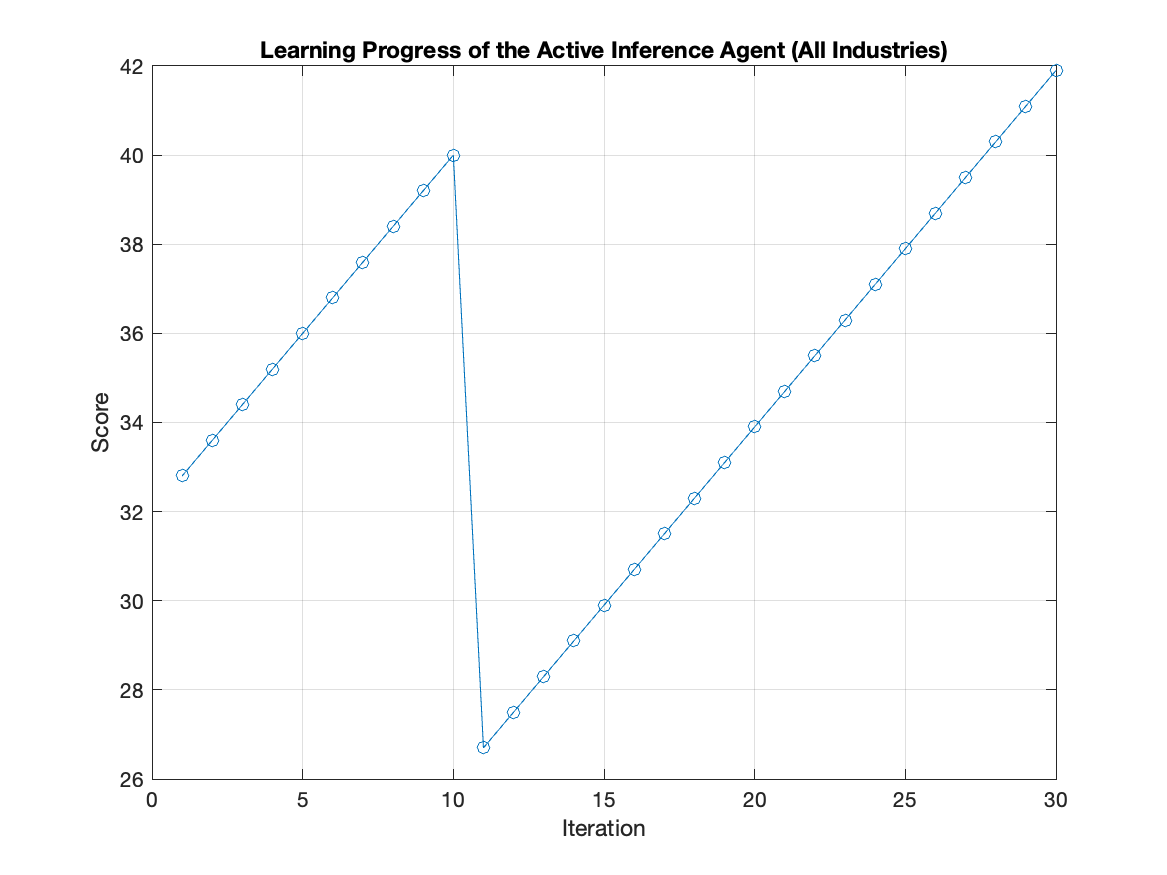}
    \caption{Learning progress of the active inference agent across all industries in the original environment. After an initial drop due to uncertainty, the agent learns the correct mappings, achieving a higher score over 20 trials.}
    \label{fig:learning_progress_all_industries_new}
\end{figure}

The first environment consists of predefined state-outcome mappings for each of the 16 industries and their respective research processes. The agent begins by interacting with the environment, learning these mappings over the first 10 trials. The first 10 trials shown in \ref{fig:learning_progress}) and \ref{fig:learning_progress_all_industries_new}) are both the calculated score per iteration for just state-observation mappings in the 1st industry and state-observation mappings for all industries, respectively.

\subsection{Second Environment: First Industry Modified}

The second 20 trials in \ref{fig:learning_progress}) were carried out by the same active inference agent placed in a new environment in which only the state-observation mappings for the first industry were changed. The mappings for the other 15 industries were the same, so only the score for Industry 1 was calculated. The steep drop-off is caused by an environment change where none of the agents' former beliefs are relevant. The agent is able to learn this new environment in a linearly increasing fashion, but doesn't reach its former maximum. 

\subsection{Third Environment: More Modifications}

The second 20 trials in \ref{fig:learning_progress_all_industries_new}) were carried out by the active inference agent initially trained in environment 1 being placed in a new environment. Here several state-observation mappings were changed across a much bigger set of industries, representing a bigger regime shift in this environment than in the second environment.

Here the agent is able to relearn the environment once again, but at a higher rate than in the second environment. In fact, the agent overtakes its previous maximum score by the 18th trial in the new environment.

\section{Conclusion}

The results from the different environments demonstrate key properties of Active Inference agents in terms of adaptation and relearning.

In the second environment, where only one industry’s mappings were changed, the agent required more iterations to relearn the altered relationships, which suggests that localized changes lead to more gradual adaptation. The slower rate of learning can be attributed to the agent's reliance on prior beliefs about the unchanged aspects of the environment, which can introduce friction when adapting to small, isolated changes. The lower learning rate can also potentially be attributed to the score calculation. Since the score was only being calculated on the 1st industry it didn't have the benefit of ballooning the score due to increased confidence from other industries.

In contrast, the third environment, where more industries experienced changes, showed faster adaptation. This can be interpreted as the agent being more flexible when faced with global environmental shifts. With multiple industries changing simultaneously, the agent can globally adjust its model, resulting in faster convergence and even surpassing the peak performance seen in the first environment. This quicker learning rate may also be due to the opposite effect potentially seen in environment two. Since, there are several unchanged mappings they're increasing scores may have ballooned the overall score more than expected.

These results suggest that the structure of the environmental changes may impact adaptation rate. The results conclusively show that the active inference agent is able to reassess its views when placed in changing environments and adapt its internal model in response. 

\section{Discussion}

This paper demonstrates the potential of Active Inference as a framework for creating self-learning agents that continuously adapt to their environment. By engaging in continual interaction with their surroundings and refining internal models, Active Inference agents are able to autonomously adjust to changes over time without requiring manual intervention. This feature makes Active Inference highly applicable in domains where adaptability is crucial, such as research, quantitative finance, and healthcare.

\subsection{Application to Quantitative Finance}

In quantitative finance, trading strategies often become obsolete due to shifting market conditions. Using an Active Inference agent, we can build a system where the agent continuously observes market indicators like asset prices, volatility, and economic factors. The state factors could include \textit{market conditions} and \textit{interest rates}, while observation modalities would encompass \textit{price movements} and \textit{volume}. 

The agent would learn and update its internal model based on real-time data, adjusting its trading strategies as market conditions evolve. The generative process simulates possible future market scenarios, enabling the agent to select policies that minimize risk and maximize returns. This adaptability makes the agent resilient to rapid market fluctuations, reducing the need for constant manual updates to trading strategies.

\subsubsection{Application to Healthcare}

In healthcare, patient data changes over time, requiring constant adaptation in treatment plans. An Active Inference agent could be applied to a clinical decision support system where state factors include \textit{patient health} and \textit{treatment options}, and observation modalities could involve \textit{test results} and \textit{symptom reports}. 

The agent would continuously update its model based on incoming patient data, refining its understanding of which treatments are most effective. By simulating different treatment outcomes, the agent can recommend the best course of action, ensuring that treatment strategies evolve as the patient’s condition changes. This continual learning leads to more personalized and effective healthcare without constant human intervention.

\subsubsection{Future Work}

Moreover, future work could explore more sophisticated generative models and processes, allowing agents not only to learn new mappings but also to discover new state factors and observation modalities. This would enable agents to move beyond pre-defined structures, learning entirely new frameworks as they interact with their environment. Additionally, integrating Active Inference with other AI paradigms, such as large language models or reinforcement learning, could yield even more adaptable and capable agents. Here, active inference agents would serve as a mechanism to increase the temporal effectiveness of other agents they can manipulate.

Active Inference offers significant potential for real-world applications that require resilience and long-term sustainability. Whether in finance, healthcare, or other fields, this framework has the ability to keep AI systems relevant and adaptive, reducing the need for human intervention and making it a key tool for the future of adaptive artificial intelligence.

\section{Data Availability}
Find all relevant code and figures at https://github.com/RPD123-byte/Demonstrating-the-Continual-Learning-Capabilities-and-Practical-Application-of-Discrete-Time-Active. Subroutines for spm-MDP-VB-X and spm-MDP-check among others can be found in the spm12 package.

\end{document}